\documentclass[11pt]{article}

\ifdefined\pdfobjcompresslevel
\fi

\usepackage{silence}
\usepackage[preprint]{acl}
\usepackage{times}
\usepackage{booktabs}
\usepackage{multirow}
\usepackage{multicol}
\usepackage{latexsym}
\usepackage{amsmath}
\usepackage{algorithm}
\usepackage{algpseudocode}
\definecolor{algblue}{RGB}{25,80,150}

\usepackage[table]{xcolor}

\usepackage{graphicx}
\usepackage{multirow}
\usepackage{amssymb}

\definecolor{groupblue}{RGB}{238,246,252}
\definecolor{oursgray}{RGB}{232,232,232}

\algrenewcommand\algorithmicindent{1.2em}
\newcommand{\ourscell}[1]{\cellcolor{oursgray}\textbf{#1}}

\usepackage[T1]{fontenc}

\usepackage[utf8]{inputenc}

\usepackage{microtype}

\usepackage[most]{tcolorbox}
\usepackage{xcolor}

\usepackage[most]{tcolorbox}
\usepackage{xcolor}

\definecolor{PromptBorder}{HTML}{065F5C}  
\definecolor{PromptBack}{HTML}{F2F9F8}      

\newtcblisting{promptbox}[1]{
    breakable,                 
    colframe=PromptBorder,     
    colback=PromptBack,        
    coltitle=white,            
    fonttitle=\bfseries,       
    title={#1},                
    arc=3pt,                   
    boxrule=1.5pt,             
    left=6pt, right=6pt, top=4pt, bottom=4pt, 
    listing only,
    listing options={
        basicstyle=\ttfamily\small,
        breaklines=true,
        breakatwhitespace=false,
        columns=fullflexible
    }
}

\definecolor{TealBorder}{HTML}{065F5C} 
\definecolor{TealBack}{HTML}{F2F9F8}   

\newtcolorbox{skillbox}[1]{
    breakable,
    colframe=TealBorder,
    colback=TealBack,
    coltitle=white,
    fonttitle=\bfseries,
    title={#1},
    arc=3pt,
    boxrule=1.5pt,
    left=6pt, right=6pt, top=4pt, bottom=4pt
}

\definecolor{caseRed}{RGB}{255,190,190}
\definecolor{caseGreen}{RGB}{0,255,60}
\definecolor{caseTeal}{RGB}{0,210,210}
\definecolor{casePurple}{RGB}{215,170,230}
\definecolor{caseYellow}{RGB}{255,245,0}

\usepackage{inconsolata}
\usepackage{enumitem}

\usepackage{graphicx}

\title{Skill-3D: Evolving Scene-Aware Skills for Agentic 3D Spatial Reasoning}

\author{%
  Haoyuan Li\textsuperscript{1},
  Zhengdong Hu\textsuperscript{2},
  Jun Wang\textsuperscript{3},
  Hehe Fan\textsuperscript{1}, and
  Yi Yang\textsuperscript{1}\thanks{Corresponding author.}\\
  \textsuperscript{1}Zhejiang University, Hangzhou, China \\
  \textsuperscript{2}University of Technology Sydney, Sydney, Australia \\ 
  \textsuperscript{3}OPPO Research Institute, Shenzhen, China \\
  Project Page: \url{https://skill-3d.github.io/}
}

\begin{document}
\maketitle

\begin{abstract}
This paper explores agentic 3D spatial understanding, \emph{i.e.,} MLLM agents performing  3D reasoning through tool use. Existing methods often misuse tools and exhibit biased tool preferences under 3D scenario, leaving the agentic paradigm with only marginal gains over non-agentic strategies. We reveal that 3D spatial reasoning tasks are heterogeneous across scenes, while these agents apply a uniform tool-use strategy to all scenes rather than selecting tools according to the specific scene and task. To address this, we propose \textbf{Skill-3D}, a framework that learns self-evolving scene-aware skills. Specifically, Skill-3D identifies the task scene and records the agent's tool-use trajectory into a \emph{Scene Memory}, where successful trajectories from similar scenes are aggregated and distilled into a reusable scene-aware skill, with failed ones attached to the skill as lessons. During training, once a similar scene recurs, the corresponding skill is injected to guide the agent, producing new trajectories whose successes and failures further refine the skill, forming a loop in which the memory and the skill library co-evolve. 
Experiments show that Skill-3D substantially improves tool utilization in 3D spatial reasoning (from {39\%} to {78\%} on VSI-Bench), driving the agent toward correct and sufficient tool use. For instance, it improves Gemini-3-Flash by {67\%} on MMSI-Bench. Furthermore, we conduct agentic post-training over skill-guided trajectories, which boosts Qwen3-VL-8B by {60\%} on VSI-Bench.
\end{abstract}

\section{Introduction}



Agentic 3D spatial reasoning aims to enable multimodal large language model (MLLM) agents to solve indoor 3D understanding tasks through external tool use, by which they can acquire spatial and geometric evidence that is difficult to infer from the MLLM alone~\cite{wu2025spatial,zhang2026think3d}. Recent methods explore this paradigm by iteratively invoking tools within a per-question reasoning loop, e.g., object detection and segmentation for 2D perception, depth estimation and 3D reconstruction for geometric grounding~\cite{zhang2026think3d, luo2026pySpatial, yuan2026boosting, ropero2026riemind}. However, these methods  often fail to realize the potential of tool use in 3D reasoning and exhibit preferences toward a few dominant tools, regardless of what each scene actually requires. As a result, adding tools to an MLLM does not improve spatial reasoning, and yields only marginal gains over non-agentic baselines under some scenarios.

\begin{figure*}
    \centering
    \includegraphics[width=\linewidth]{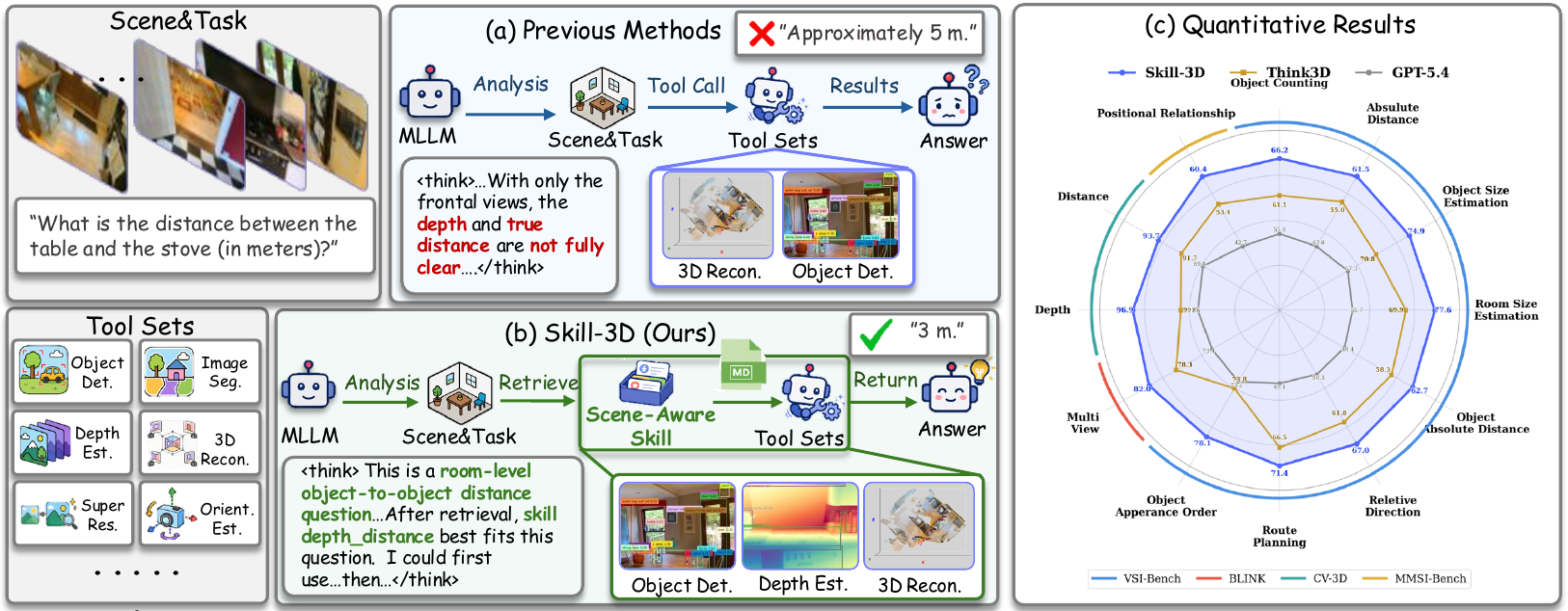}
    \caption{\textbf{Motivation and overview of Skill-3D.} \textbf{(a)} Scene-agnostic tool calls can yield mismatched evidence and unreliable answers. \textbf{(b)} Skill-3D retrieves scene-aware skills to guide tool-use workflows, e.g., detection, depth, 3D reconstruction. \textbf{(c)} Skill-3D improves over strong MLLM baselines across diverse spatial reasoning dimensions.}
    \label{fig:motivation}
    \vspace{-1.5em}
\end{figure*}

We attribute this limitation to the scene heterogeneity of indoor 3D reasoning, where required evidence and tool workflows vary across scenes. As shown in Fig.~\ref{fig:motivation}(a), the ``object-to-object distance estimation'' question requires depth evidence. However, existing methods often adopt a uniform tool strategy and rely on object detection and 3D reconstruction, which mainly provide relative spatial relationships rather than the depth grounding needed for absolute distance estimation. Sec.~\ref{sec:ablation_study} confirms that this failure consistently occurs across diverse 3D scenes.

In this work, we propose \textbf{Skill-3D}, a framework that equips MLLM agents with reusable scene-aware skills. As illustrated in Fig.~\ref{fig:motivation}(b), given the same ``object-to-object distance estimation'' question, Skill-3D identifies the scene-task context, retrieves a relevant skill, and invokes suitable perception tools such as depth estimation. It learns these skills by constructing a Scene Memory and co-evolving a Skill Library on top of it.

During training, an MLLM agent identifies each question's scene and stores the corresponding tool-use trajectory together with its outcome into the Scene Memory. On top of this memory, the Skill Library aggregates successful trajectories from similar scenes and distills them into reusable scene-aware skills, with failed ones attached to the corresponding skill as lessons. Critically, once a skill is formed, it is injected back to guide the agent on subsequent questions from similar scenes, producing new trajectories whose successes and failures are updated back to refine the same skill. Through this loop, the Scene Memory and the Skill Library co-evolve until the skills are reliable enough to serve as scene-conditioned tool-use priors at inference.



This design offers two practical benefits. \textbf{1)} Skills are dynamically updated: under a similar scene, the agent's new trajectories are written back to broaden the skill's coverage. This prevents the skill from overfitting to a narrow slice of its scene (\emph{e.g.,} kitchen depth-estimation vs. living room depth-estimation). \textbf{2)} The Scene Memory and the Skill Library evolve together, with neither predefined upfront, allowing both to become more discriminative as the agent encounters more diverse 3D tasks.


Additionally, we further introduce \textbf{skill-guided agentic post-training}. We first apply supervised fine-tuning on skill-guided trajectories to teach the policy the format of skill retrieval, tool invocation, and evidence accumulation. We then perform Group Relative Policy Optimization (GRPO)~\cite{DeepSeekAI2025DeepSeekR1IR, shao2024deepseekmath} with a composite reward that jointly captures answer correctness, skill-guided tool-use quality, and structured output, encouraging the policy to internalize the scene-aware tool-use behavior that the skill library encodes.

We evaluate Skill-3D on multiple 3D spatial reasoning benchmarks. As shown in Fig.~\ref{fig:motivation}(c), Skill-3D consistently outperforms strong MLLM baselines across representative 3D reasoning dimensions, improving effective tool usage from {39\%} to {78\%}. It lifts Gemini-3-Flash by {67\%} on MMSI-Bench, while skill-guided agentic post-training further boosts Qwen3-VL-8B~\cite{Qwen3-VL} by {43\%} on VSI-Bench~\cite{yang2025thinking}. Our contributions are threefold:
\begin{itemize}[leftmargin=*, itemsep=0pt, topsep=2pt]
\item  We propose {Skill-3D}, which constructs a Scene Memory and co-evolves a Skill Library on top of it during training, yielding scene-aware skills that generalize across scene-internal variations.

\item We propose skill-guided agentic reinforcement learning under a composite reward, internalizing scene-aware tool-use behavior into the policy.

\item Extensive experiments across closed- and open-source MLLMs on multiple 3D spatial reasoning benchmarks validate the effectiveness of Skill-3D and its substantial improvement in tool usage.
\end{itemize}

\section{Related Work}

\subsection{MLLMs for Spatial Reasoning}
Multimodal Large Language Models (MLLMs) have shown growing capability in spatial reasoning, driven by stronger backbones~\cite{yang2023mm, wake2024gpt, shao2024visual, liu2025coarse, lee2025perspective} and dedicated benchmarks~\cite{yang2025thinking, wu2025spatialscore, chow2025physbench, cai2025spatialbot, majumdar2024openeqa, liu2026openspatial, zhang2026revsi}. Recent methods improve fine-grained spatial understanding by incorporating 3D reconstruction, depth cues, spatial VQA data, and explicit grounding~\cite{cheng2024spatialrgpt, chen2024spatialvlm, fan2025vlm, roy2025bydeway, GPT4Scene, huang2024chat, wang2023chat, balazadeh2024synthetic, zhang2025spatial, wu2025reinforcing}. Other works enhance spatial reasoning through prompting, mental simulation, visual chain-of-thought, reinforcement learning, code-driven 3D reasoning, and generative imagination of 3D space~\cite{taguchi2025spatialprompting, marsili2025visual, tang2025video, lee2025perspective, fan2025grit, wang2025visuothink, wang2025perception, chen2025geometrically, luo2026pySpatial, yang2025mindjourney}. These capabilities have also been extended to embodied and robotic settings~\cite{ji2025robobrain, team2025robobrain, team2025gemini, abdolmaleki2025gemini, zhou2025roborefer, zhou2024navgpt, zhao2026cov}.

\subsection{MLLM Agents}
Tool augmentation extends MLLM by allowing them to invoke external modules through prompting, structured APIs, or code generation. Representative systems demonstrate that external tools can compensate for limitations of end-to-end multimodal models~\cite{shen2023hugginggpt, wu2023visual, suris2023vipergpt}. Recent tool-augmented VLM agents have been developed for long-video understanding, high-resolution image analysis, medical diagnosis, and general visual reasoning~\cite{chen2025lvagent, zhang2025deep, taguchi2025spatialprompting, yang2025vca, zhu2025segagent, lee2025training, yang2025visionthink, lyu2025wsi, liu2025insightx, su2025openthinkimg}. A complementary line of work trains VLMs to use tools through supervised fine-tuning or reinforcement learning~\cite{liu2024llava, wang2025mllm, han2025tiger, tang2025can, wu2024dettoolchain, lin2025olympus, wu2025vtool, zheng2025driveagent, chen2025learning, dong2025agentic, zhou2025reinforced}. Recent 3D agentic methods further introduce reconstruction-based reasoning loops for limited-view spatial understanding~\cite{zhang2026think3d}, but they often rely on uniform tool-use workflows across heterogeneous scenes.

\subsection{Agent Skills}
Memory-based agents store trajectories for reflection or experience replay~\cite{zhao2024expel, shinn2024reflexion}, but raw trajectories are often long, redundant, and noisy~\cite{chhikara2025mem0, yan2025memory}. Recent work therefore studies \textit{skills}: reusable behavioral primitives distilled from historical interactions~\cite{xu2026agentskillsurvey, li2026skillecosystem, he2026openclaw, yang2026skilloptexecutivestrategyselfevolving}. Skills can serve as procedural memory for decision-time guidance~\cite{li2026skillsbench, liu2026self-vla, liang2026skillnet, jiang2026xskill, zhang2026memskill, ye2026meta} and can also provide high-level priors for reinforcement learning~\cite{xia2026skillrl, wang2025rlskill, jiao2026agenticproposing, ouyang2026skillos, fan2026exploring}. Existing skill-based agents mainly study general task automation, skill retrieval, or policy improvement. In contrast, Skill-3D studies skills for 3D spatial reasoning, where skills must encode perception-grounded tool workflows involving objects, geometry, and multi-view evidence. 
\begin{figure*}[t]
    \centering
    \includegraphics[width=\linewidth]{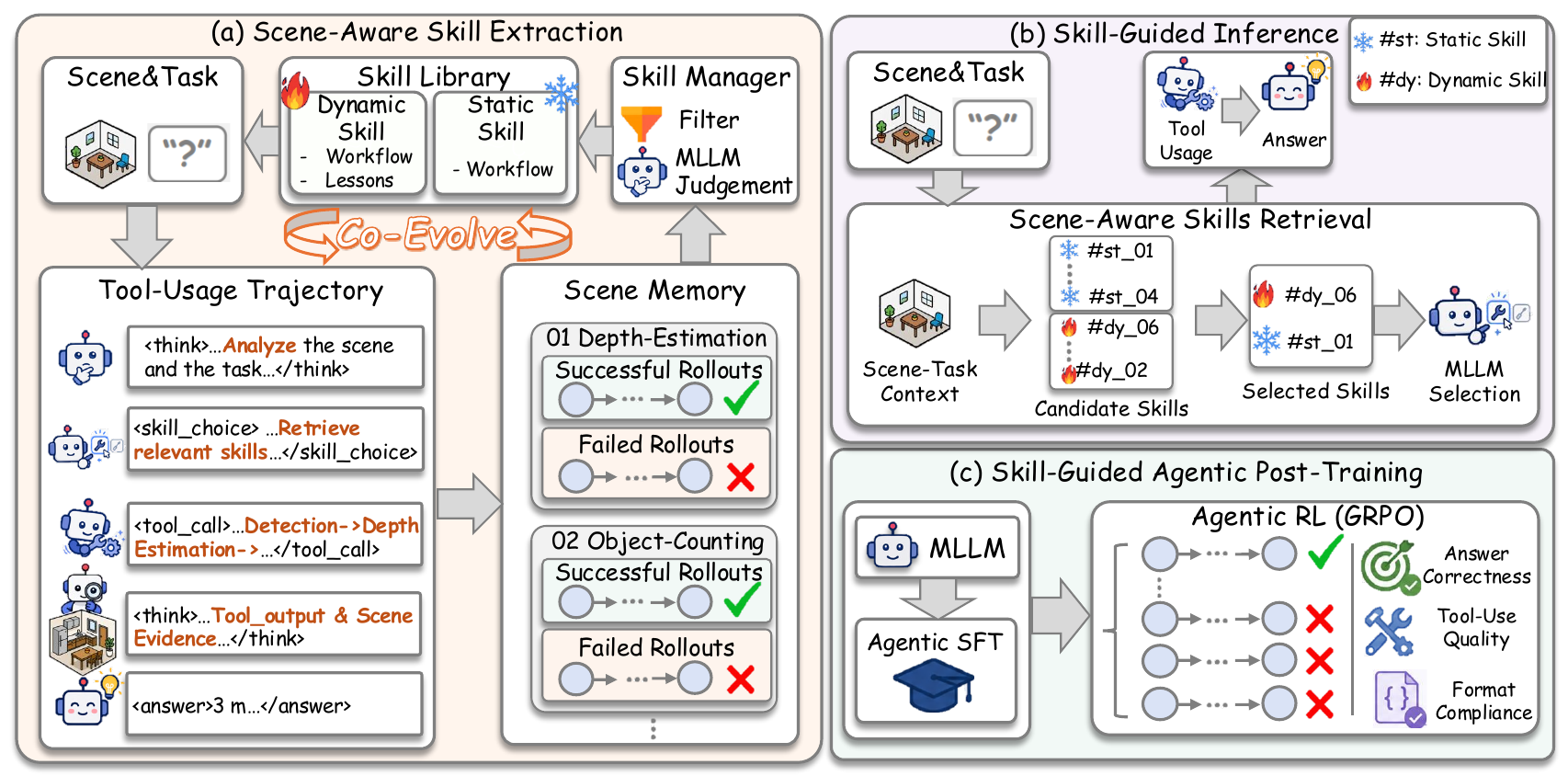}
    \caption{
    \textbf{Overview of Skill-3D.}
    \textbf{(a)} Skill-3D records scene-task rollouts into Scene Memory, which stores scene context, tool evidence, and failure patterns. Successful rollouts are distilled into dynamic skills, while failed rollouts are attached as lessons, enabling Scene Memory and the Skill Library to co-evolve. \textbf{(b)} Given a new query, Skill-3D identifies the scene-task context, retrieves relevant static and dynamic skills, and selects a compact skill set to guide tool-use workflow and evidence acquisition. \textbf{(c)} Skill-guided trajectories are used for agentic SFT and GRPO, encouraging compact agents to internalize skill selection, tool use, and evidence-grounded spatial reasoning.
    }
    \label{fig:method}
    \vspace{-1.5em}
\end{figure*}

\section{Method}
\label{sec:method}

In this section, we present \textbf{Skill-3D}, a scene-aware skill learning framework for agentic 3D spatial reasoning. As shown in Fig.~\ref{fig:method}, Skill-3D consists of three stages. First, it records completed rollouts into Scene Memory and evolves a Skill Library from both successes and failures. The Scene Memory stores rollouts collected across benchmarks, allowing dynamic skills to be formed from heterogeneous spatial reasoning cases rather than being restricted to a single benchmark. Second, Skill-3D retrieves scene-task-relevant skills to guide inference-time tool-use planning. Third, it uses skill-guided trajectories to post-train compact agents through agentic Supervised Fine-Tuning (SFT) and Reinforcement Learning (RL).

\subsection{Scene-Aware Skill Extraction}

Given a spatial question $q$ and a set of visual observations $O=\{o_i\}_{i=1}^{N}$ from an indoor scene, an MLLM agent predicts an answer $\hat{y}$ by optionally invoking tools from tool sets $\mathcal{T}$. The tool sets include external perception and geometry modules, e.g., object detection, segmentation, depth estimation, orientation estimation, super-resolution, and 3D reconstruction. A rollout contains the question, observations, reasoning trace, selected skills, tool calls, tool outputs, and final answer. Skill-3D updates the Skill Library after each completed rollout. Successful rollouts provide reusable tool-use patterns, while failed rollouts provide diagnostic signals for future correction.

\noindent\textbf{Successes as Workflows.}
For each successful rollout, Skill-3D extracts a reusable tool-use routine, including its trigger condition, required evidence, tool order, key arguments, and evidence-to-answer mapping. The routine is promoted to a new dynamic skill if no compatible skill exists; otherwise, it is merged into an existing skill only when it adds useful coverage, such as a new scene condition, stronger evidence source, or lower-cost workflow. If it provides no new information, Skill-3D only updates the success statistics of the matched skill. This keeps the Skill Library compact while expanding the coverage of existing skills.

\noindent\textbf{Failures as Lessons.}
Failed rollouts are not discarded. Skill-3D diagnoses each failure from its Scene Context and Tool Usage, with typical error types including wrong tool selection, missing evidence, invalid tool input, ignored tool output, and redundant tool calls. Evidence-supported failures are attached to the related skill as lessons. When a failure suggests a reliable correction, the corresponding dynamic skill is patched with a fallback rule. When similar failures repeatedly occur under a static skill, Skill-3D creates a failure-aware dynamic skill to handle that recurring case.

\noindent\textbf{Skill Maintenance.}
The Skill Manager keeps the active Skill Library compact and reliable by filtering noisy rollouts and deciding whether each candidate update should be inserted, merged, patched, or rejected. An update is accepted only when it is evidence-supported and consistent with previous successful cases. Successful updates are promoted to new dynamic skills or merged into compatible ones, while failure updates are attached as lessons or converted into fallback rules. Static skills remain fixed as task-level priors, whereas dynamic skills evolve through validated merges and patches. Thus, the library stores reusable scene-aware procedures rather than raw trajectories. Please see Appendix~\ref{app:prompt_design} for detailed prompt design.

\subsection{Skill-Guided Inference}
\label{sec:skill_guided_inference}

Given a new query, Skill-3D first identifies the scene-task context, including the task category, target entities, scene signature, and required evidence. This context determines whether the agent should seek object-level evidence, boundary evidence, depth cues, orientation cues, multi-view geometry, or a combination of them.

\noindent\textbf{Scene-Task Skill Retrieval.}
Skill-3D performs top-$k$ retrieval over the Skill Library to obtain candidate static and dynamic skills. Each skill is indexed by its trigger condition, applicable scene context, required evidence type, and historical metadata. Given the current scene-task context, we score each skill by its semantic alignment with the query category, target entities, scene signature, and evidence requirement, e.g., whether the task requires object boundaries, depth cues, orientation evidence, or multi-view geometry. The ranking also incorporates metadata including historical success rate, attached failure lessons, and estimated tool cost. This retrieval step returns a compact set of potentially useful skills without injecting the entire Skill Library into the prompt.

\noindent\textbf{Skill Selection.}
The candidate skills may contain redundant or overlapping workflows. Skill-3D therefore uses the policy to select a compact subset of skills for the current query. The selected skills are expected to cover the required evidence while avoiding unnecessary tool calls. The selector also generates short fallback rules, e.g., switching from detection to segmentation when closest-point boundaries are required, or using multi-view evidence when single-view localization is ambiguous.

\noindent\textbf{Tool-Use Workflow.}
Conditioned on the selected skill, the agent performs iterative tool reasoning. At each step, the model decides whether to invoke a tool, incorporate returned evidence, continue reasoning, or stop and answer. Tool outputs are appended to the reasoning history and used to update the accumulated evidence. Compared with direct tool invocation, skill-guided tool-use workflow constrains both evidence acquisition and evidence usage. The agent is guided to collect the evidence required by the scene-task context and to ground the final answer in the returned tool outputs.

\subsection{Skill-Guided Agentic Post-Training}

Skill-3D further transfers scene-aware tool-use behavior into compact MLLM agents. During agentic post-training, the Skill Library is frozen to avoid non-stationarity. Each training sample contains the question and observations, available skill candidates, the selected skill sequence, tool calls and outputs, intermediate evidence, and final answer.

\noindent\textbf{Agentic SFT.}
We first perform SFT on skill-guided trajectories. This stage teaches the model the complete structured interaction pattern. Importantly, the SFT target is not only to imitate tool calls, but also to learn when and how to select suitable skills from the Skill Library according to the scene-task context. This provides a stable initialization so that the policy can execute skill selection, tool-use workflow, and evidence integration before RL.

\noindent\textbf{Agentic RL.}
We further optimize the skill-augmented policy with Group Relative Policy Optimization (GRPO)~\cite{DeepSeekAI2025DeepSeekR1IR, shao2024deepseekmath}. For each scene-task query, the policy first observes the question, visual observations, and retrieved skill candidates. It then samples a group of $G$ complete trajectories
$\{\tau^{(1)},\ldots,\tau^{(G)}\}$, where each trajectory contains the model's own skill choices, tool calls, tool outputs, reasoning steps, and final answer. Each trajectory receives a scalar reward:
\begin{equation}
R(\tau)=R_{\mathrm{ans}}(\tau)
+R_{\mathrm{fmt}}(\tau)+R_{\mathrm{tool}}(\tau),
\label{eq:post_training_reward}
\end{equation}
where $R_{\mathrm{ans}}$ measures answer correctness, $R_{\mathrm{fmt}}$ measures structured-format compliance, and $R_{\mathrm{tool}}$ measures tool-use efficiency, i.e., whether the selected tools provide useful evidence with minimal redundant calls. Specifically, we define $R_{\mathrm{tool}}$ as:
\begin{equation}
R_{\mathrm{tool}}(\tau)
=
R_{\mathrm{exec}}(\tau)
-
\frac{|\mathcal{A}|}{B},
\end{equation}
where $\mathcal{A}$ is the set of tool calls in trajectory $\tau$, $B$ is the maximum tool budget. In practice, $R_{\mathrm{exec}}$ is a binary reward assigned to 1 only when the trajectory obtains the required evidence specified by the benchmark task type and the frozen scene-task parser. The required evidence is not determined by the model-selected skill, which prevents the policy from selecting easier skills to obtain higher tool-use reward. We provide additional objective details in Appendix~\ref{app:rl}.


\begin{table*}[!t]
\centering
\captionof{table}{\textbf{Comprehensive evaluation on VSI-Bench, BLINK, CV-3D, and MMSI-Bench.} We report representative spatial reasoning metrics across multiple benchmarks. MV
denotes multi-view. PR denotes positional relationship. Higher values indicate better performance. All reported metrics are obtained on the test set.}
\setlength\tabcolsep{1.8pt}
\renewcommand{\arraystretch}{1.03}
\resizebox{\textwidth}{!}{
\begin{tabular}{c|l|cccccccc|c|cc|c}
\toprule
\multirow{2}{*}{\textbf{Model}} & \multirow{2}{*}{\textbf{Method}} 
& \multicolumn{8}{c|}{\textbf{VSI}} 
& \multicolumn{1}{c|}{\textbf{BLINK}} 
& \multicolumn{2}{c|}{\textbf{CV-3D}} 
& \multicolumn{1}{c}{\textbf{MMSI}} \\
\cmidrule(lr){3-10}\cmidrule(lr){11-11}\cmidrule(lr){12-13}\cmidrule(lr){14-14}
& & Obj. Cnt. & Abs. Dist. & Obj. Size & Room Size & Rel. Dist. & Rel. Dir. & Route Plan & Appr. Order
& MV
& Depth Order & Rel. Dist.
& PR \\

\midrule

\multirow{4}{*}{\rotatebox[origin=c]{90}{\small GPT-4o}}
& w/o Tools 
& 38.1 & 7.9 & 29.6 & 37.4 & 36.5 & 35.6 & 26.3 & 28.2
& 47.9
& 72.6 & 70.9
& 28.7 \\
& w/ Tools
& 48.7 & 12.1 & 47.0 & 40.9 & 41.6 & 44.0 & 34.3 & 29.7
& 60.8
& 86.6 & 84.9
& 34.6 \\
& Think3D
& 50.4 & 32.7 & 50.5 & 61.4 & 47.9 & 52.6 & 55.9 & 29.2
& 62.7
& 88.3 & 86.1
& 38.4 \\
& \ourscell{Skill-3D}
& \ourscell{56.8} & \ourscell{42.6} & \ourscell{58.1} & \ourscell{69.5} & \ourscell{53.4} & \ourscell{59.2} & \ourscell{62.7} & \ourscell{35.4}
& \ourscell{72.4}
& \ourscell{92.0} & \ourscell{90.5}
& \ourscell{43.2} \\

\midrule

\multirow{4}{*}{\rotatebox[origin=c]{90}{\small GPT-5.4}}
& w/o Tools
& 55.8 & 43.6 & 67.3 & 55.7 & 48.4 & 50.5 & 49.1 & 73.2
& 73.4
& 91.6 & 89.8
& 42.7 \\
& w/ Tools
& 58.0 & 47.8 & 69.3 & 58.2 & 51.7 & 53.3 & 52.5 & 74.6
& 75.1
& 92.5 & 90.1
& 48.1 \\
& Think3D
& 61.1 & 55.0 & 70.8 & 69.9 & 58.3 & 61.8 & 66.5 & 73.8
& 78.3
& 93.0 & 91.7
& 53.4 \\
& \ourscell{Skill-3D}
& \ourscell{66.2} & \ourscell{61.5} & \ourscell{74.9} & \ourscell{77.6} & \ourscell{62.7} & \ourscell{67.0} & \ourscell{71.4} & \ourscell{78.1}
& \ourscell{82.0}
& \ourscell{96.9} & \ourscell{93.7}
& \ourscell{60.4} \\

\midrule

\multirow{4}{*}{\rotatebox[origin=c]{90}{\small Gemini-2.5-Pro}}
& w/o Tools
& 45.9 & 37.6 & 62.2 & 42.8 & 60.5 & 45.9 & 42.7 & 70.4
& 70.6
& 90.7 & 90.3
& 36.9 \\
& w/ Tools
& 48.4 & 41.5 & 66.0 & 46.3 & 62.6 & 51.2 & 49.5 & 72.7
& 72.3
& 91.4 & 91.2
& 44.3 \\
& Think3D
& 58.2 & 53.1 & 69.5 & 66.4 & 64.8 & 59.5 & 65.8 & 72.2
& 76.0
& 92.7 & 91.6
& 51.0 \\
& \ourscell{Skill-3D}
& \ourscell{62.4} & \ourscell{58.0} & \ourscell{73.1} & \ourscell{72.8} & \ourscell{67.6} & \ourscell{64.2} & \ourscell{69.0} & \ourscell{76.4}
& \ourscell{79.2}
& \ourscell{94.0} & \ourscell{92.8}
& \ourscell{56.7} \\

\midrule

\multirow{4}{*}{\rotatebox[origin=c]{90}{\small Gemini-3-Flash}}
& w/o Tools
& 45.3 & 9.2 & 45.7 & 39.8 & 38.7 & 42.2 & 33.8 & 31.3
& 59.1
& 84.6 & 82.8
& 32.7 \\
& w/ Tools
& 48.2 & 13.7 & 48.8 & 42.1 & 42.3 & 44.2 & 36.4 & 32.8
& 61.3
& 86.2 & 83.7
& 36.8 \\
& Think3D
& 56.8 & 52.3 & 68.0 & 66.3 & 56.5 & 60.8 & 64.2 & 69.3
& 75.2
& 91.8 & 91.0
& 49.2 \\
& \ourscell{Skill-3D}
& \ourscell{60.9} & \ourscell{56.1} & \ourscell{71.2} & \ourscell{71.8} & \ourscell{60.4} & \ourscell{63.0} & \ourscell{67.5} & \ourscell{73.4}
& \ourscell{77.6}
& \ourscell{93.2} & \ourscell{92.1}
& \ourscell{54.8} \\

\bottomrule
\end{tabular}
}
\vspace{-1em}
\label{tab:comprehensive_benchmarks}
\end{table*}
\section{Experiments}
\label{sec:experiments}


\subsection{Experimental Setup}

\noindent\textbf{Benchmarks and Metrics.}
We evaluate on VSI-Bench~\cite{yang2025thinking}, BLINK~\cite{fu2024blink}, CV-3D~\cite{tong2024cambrian}, and MMSI-Bench~\cite{yang2025mmsi}. VSI-Bench covers eight indoor spatial reasoning categories, including object counting, distance estimation, size estimation, route planning, and appearance order. BLINK evaluates multi-view reasoning, CV-3D evaluates depth ordering and relative distance, and MMSI-Bench evaluates positional relationship reasoning. As shown in Table~\ref{tab:dataset_split}, we use the scripts provided by Think3D~\cite{zhang2026think3d} to randomly sample 30\% of the questions from each category in each benchmark as the training set, and use the remaining disjoint samples as the test set for fair comparison. For VSI-Bench, we follow Think3D~\cite{zhang2026think3d} and uniformly sample seven frames from the full scene video to serve as model input. More details about benchmarks and metrics can be found in Appendix~\ref{app:dataset_details}.

\noindent\textbf{Models and Baselines.}
For closed-source agents, we evaluate GPT-4o~\cite{hurst2024gpt}, GPT-5.4 ~\cite{openai2025introducing}, Gemini-2.5-Pro~\cite{comanici2025gemini}, and Gemini-3-Flash~\cite{google2025gemini3}. Each backbone is tested under four settings: w/o Tools, w/ Tools, Think3D~\cite{zhang2026think3d}, and Skill-3D. For open-source agents, we evaluate Qwen3-VL-4B~\cite{Qwen3-VL} and Qwen3-VL-8B~\cite{Qwen3-VL} with the same settings, where Skill-3D-4B and Skill-3D-8B denote skill-guided post-trained models.

\noindent\textbf{Implementation details.}
Skill-3D uses external tools to help
agents understand 3D scenes, e.g., Pi3~\cite{wang2025pi}, GroundingDINO~\cite{liu2024grounding}, SAM3~\cite{carion2025sam}, Orient Anything v2~\cite{wang2026orient}, SwinIR~\cite{liang2021swinir}, and the indoor metric-depth variant of Depth Anything 3~\cite{lin2025depth}. We use Qwen3-VL-4B/8B as our base model and GPT-5.4 as the teacher model for skill distillation and SFT data generation. The teacher model is used only on training samples for skill distillation and SFT data generation. The training set contains 500 samples for SFT and 1k samples for GRPO. We train for one epoch with a composite reward consisting of answer correctness, tool-use efficiency and skill-tool format rewards with weights 0.6, 0.2 and 0.2, respectively. We conduct experiments on 4 NVIDIA RTX PRO 6000 Blackwell GPUs. We construct a single global Scene Memory and Skill Library by pooling the training splits of all benchmarks, and freeze the resulting library during evaluation and post-training. The SFT stage takes approximately 3 hours. The RL stage takes approximately 28 hours. More details can be found in Appendix~\ref{app:hyperparameters}.

\begin{table*}[!t]
\centering
\captionof{table}{\textbf{Open-source evaluation on VSI-Bench, BLINK, CV-3D, and MMSI-Bench.} We report representative spatial reasoning metrics across multiple benchmarks. MV denotes multi-view. PR denotes positional relationship. Higher values indicate better performance. All reported metrics are obtained on the test set.}
\setlength\tabcolsep{1.8pt}
\renewcommand{\arraystretch}{1.05}
\resizebox{\textwidth}{!}{
\begin{tabular}{c|l|cccccccc|c|cc|c}
\toprule
\multirow{2}{*}{\textbf{Model}} & \multirow{2}{*}{\textbf{Method}} 
& \multicolumn{8}{c|}{\textbf{VSI}} 
& \multicolumn{1}{c|}{\textbf{BLINK}} 
& \multicolumn{2}{c|}{\textbf{CV-3D}} 
& \multicolumn{1}{c}{\textbf{MMSI}} \\
\cmidrule(lr){3-10}\cmidrule(lr){11-11}\cmidrule(lr){12-13}\cmidrule(lr){14-14}
& & Obj. Cnt. & Abs. Dist. & Obj. Size & Room Size & Rel. Dist. & Rel. Dir. & Route Plan & Appr. Order
& MV
& Depth Order & Rel. Dist.
& PR \\
\midrule

\multirow{4}{*}{\rotatebox[origin=c]{90}{\small Qwen3-VL-4B}}
& w/o Tools
& 26.6 & 19.1 & 22.8 & 35.2 & 34.8 & 35.0 & 29.4 & 40.1
& 35.6
& 59.7 & 58.6
& 26.3 \\
& w/ Tools
& 38.4 & 23.7 & 41.2 & 39.3 & 37.6 & 42.8 & 36.5 & 45.3
& 48.5
& 73.9 & 72.3
& 31.4 \\
& Think3D-4B
& 41.5 & 29.4 & 44.2 & 48.7 & 29.6 & 44.1 & 30.8 & 52.2
& 48.7
& 75.3 & 73.4
& 33.8 \\
& \ourscell{Skill-3D-4B}
& \ourscell{48.6} & \ourscell{36.8} & \ourscell{50.2} & \ourscell{57.4} & \ourscell{43.5} & \ourscell{50.4} & \ourscell{48.8} & \ourscell{56.7}
& \ourscell{60.8}
& \ourscell{79.0} & \ourscell{77.2}
& \ourscell{38.2} \\

\midrule

\multirow{4}{*}{\rotatebox[origin=c]{90}{\small Qwen3-VL-8B}}
& w/o Tools
& 32.5 & 24.3 & 30.8 & 41.6 & 40.2 & 41.6 & 35.5 & 47.3
& 43.8
& 68.8 & 66.5
& 31.0 \\
& w/ Tools
& 44.7 & 27.8 & 48.1 & 46.0 & 44.7 & 48.4 & 43.2 & 52.6
& 57.4
& 82.9 & 80.7
& 36.6 \\
& Think3D-8B
& 48.3 & 38.5 & 51.2 & 58.1 & 41.6 & 52.9 & 45.4 & 60.8
& 61.7
& 85.0 & 83.3
& 41.2 \\
& \ourscell{Skill-3D-8B}
& \ourscell{56.5} & \ourscell{48.6} & \ourscell{59.8} & \ourscell{67.9} & \ourscell{52.0} & \ourscell{60.1} & \ourscell{58.4} & \ourscell{66.8}
& \ourscell{68.5}
& \ourscell{89.6} & \ourscell{87.4}
& \ourscell{42.8} \\

\bottomrule
\end{tabular}
}
\vspace{-1em}
\label{tab:open_source_comprehensive}
\end{table*}
\subsection{Main Results}

\noindent\textbf{Closed-Source Agents.}
Table~\ref{tab:comprehensive_benchmarks} shows that Skill-3D consistently outperforms non-agentic, direct tool-use, and Think3D baselines across all four closed-source MLLM agents. Unlike benchmark-specific memories, Skill-3D uses a single shared Skill Library constructed from heterogeneous spatial reasoning benchmarks, allowing reusable skills learned from one benchmark to transfer to other benchmarks when similar scene-task contexts are encountered. The gains are most pronounced on VSI-Bench, where diverse tasks such as counting, distance estimation, size estimation, direction reasoning, and route planning require different evidence sources. Averaged over the four closed-source agents, Skill-3D improves the VSI-Bench average from 42.9 to 64.5, corresponding to a 50.3\% relative gain over the w/o Tools baseline. Consistent improvements are also observed on BLINK, CV-3D, and MMSI-Bench. Compared with direct tool use and Think3D, these results suggest that Skill-3D benefits not merely from tool availability or generic 3D reconstruction, but from retrieving scene-aware skills that specify task-relevant evidence and tool workflows. Qualitative results are shown in Fig.~\ref{fig:visualize1} in Appendix~\ref{sec:qualitative_results}.

\noindent\textbf{Open-Source Agents.}
Table~\ref{tab:open_source_comprehensive} further shows that Skill-3D transfers effectively to compact open-source agents. On VSI-Bench, Skill-3D-4B achieves a 59.7\% relative gain over the w/o Tools baseline, while Skill-3D-8B achieves a 60.3\% relative gain. These improvements are achieved across diverse spatial reasoning categories, indicating that the learned skills are not tied to a single task type. The stronger 8B results indicate that larger base models can better exploit retrieved skills and tool evidence, while the consistent 4B improvements show that scene-aware skills can still be learned by smaller agents through post-training. Qualitative results are shown in Fig.~\ref{fig:visualize2} in Appendix~\ref{sec:qualitative_results}.

\subsection{Ablation Study}
\label{sec:ablation_study}
\begin{figure}
    \centering
    \includegraphics[width=\linewidth]{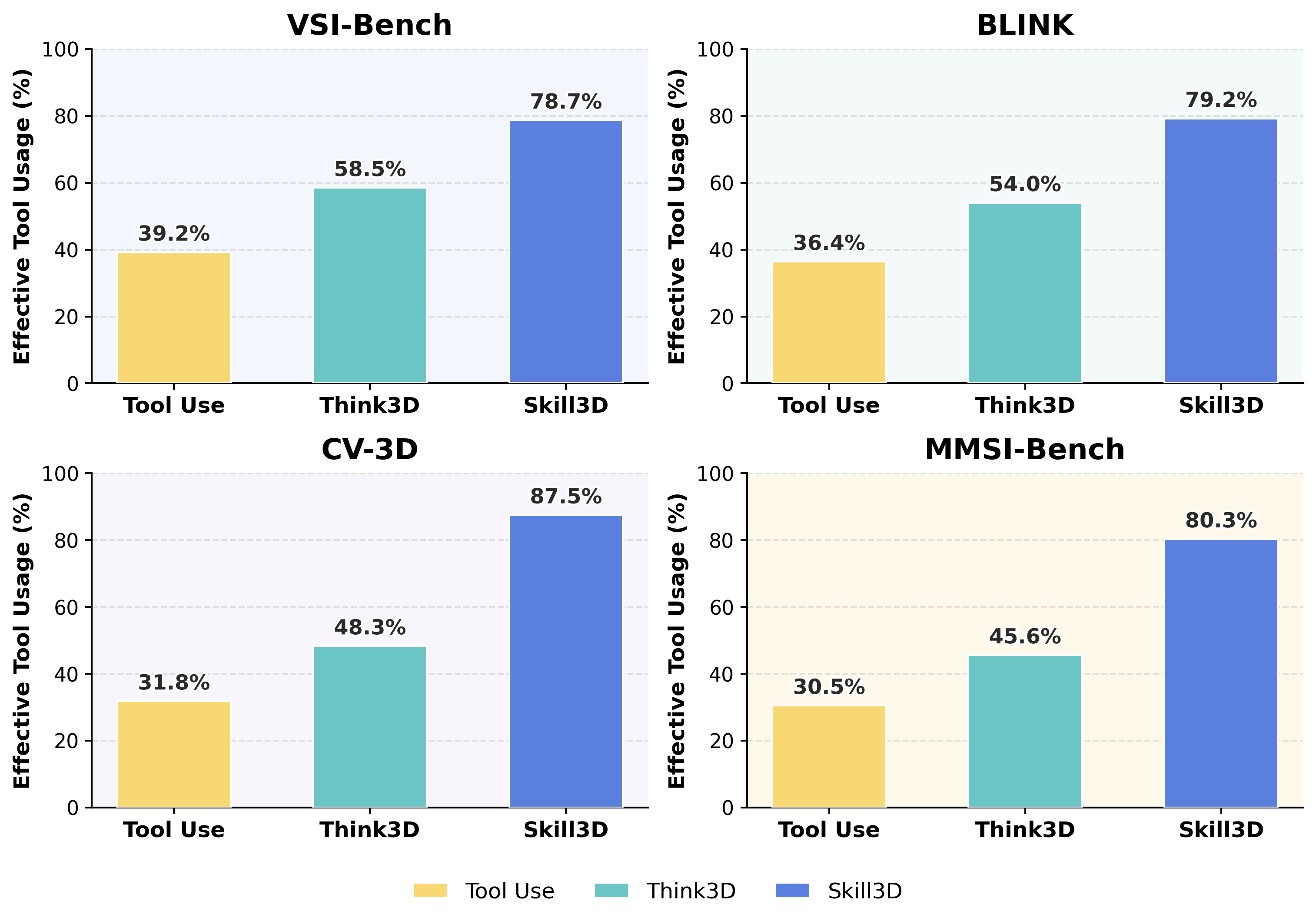}
    \caption{\textbf{Effective tool usage analysis.} We report the percentage of tool calls that contribute valid, relevant evidence to the final answer across VSI-Bench, BLINK, CV-3D, and MMSI-Bench.}
    \label{fig:tool_usage}
    \vspace{-1.5em}
\end{figure}

\noindent\textbf{Effective Tool Usage.}
We further examine whether Skill-3D improves tool-use quality rather than merely increasing the number of tool calls. We report effective tool usage (ETU), which measures the fraction of invoked tools that return valid evidence and are actually used by the agent:
\begin{equation}
\mathrm{ETU} =
\frac{1}{|\mathcal{A}|}
\sum_{a\in\mathcal{A}}
\mathbb{I}\left[\mathrm{Valid}(a)\land \mathrm{Used}(a)\right],
\end{equation}
where $\mathcal{A}$ denotes the set of tool calls in a completed rollout. $\mathrm{Valid}(a)$ is determined from tool execution logs and indicates that tool call $a$ returns non-empty and usable evidence. $\mathrm{Used}(a)$ indicates that the returned evidence is substantively consumed in the subsequent workflow, i.e., it is referenced in later reasoning, passed to downstream tools, or used to support the final answer. We compute ETU after the full rollout is completed, so that both tool execution validity and downstream evidence usage can be assessed. As shown in Fig.~\ref{fig:tool_usage}, Skill-3D substantially improves ETU over the direct Tool-Use setting, from 39.2\% to 78.7\% on VSI-Bench, 36.4\% to 79.2\% on BLINK, 31.8\% to 87.5\% on CV-3D, and 30.5\% to 80.3\% on MMSI-Bench. Since ETU is normalized by the total number of tool calls, these gains show that Skill-3D does not simply invoke more tools. Instead, it guides the agent to select evidence-producing tools and integrate the returned evidence into spatial reasoning.

\begin{figure}
    \centering
    \includegraphics[width=\linewidth]{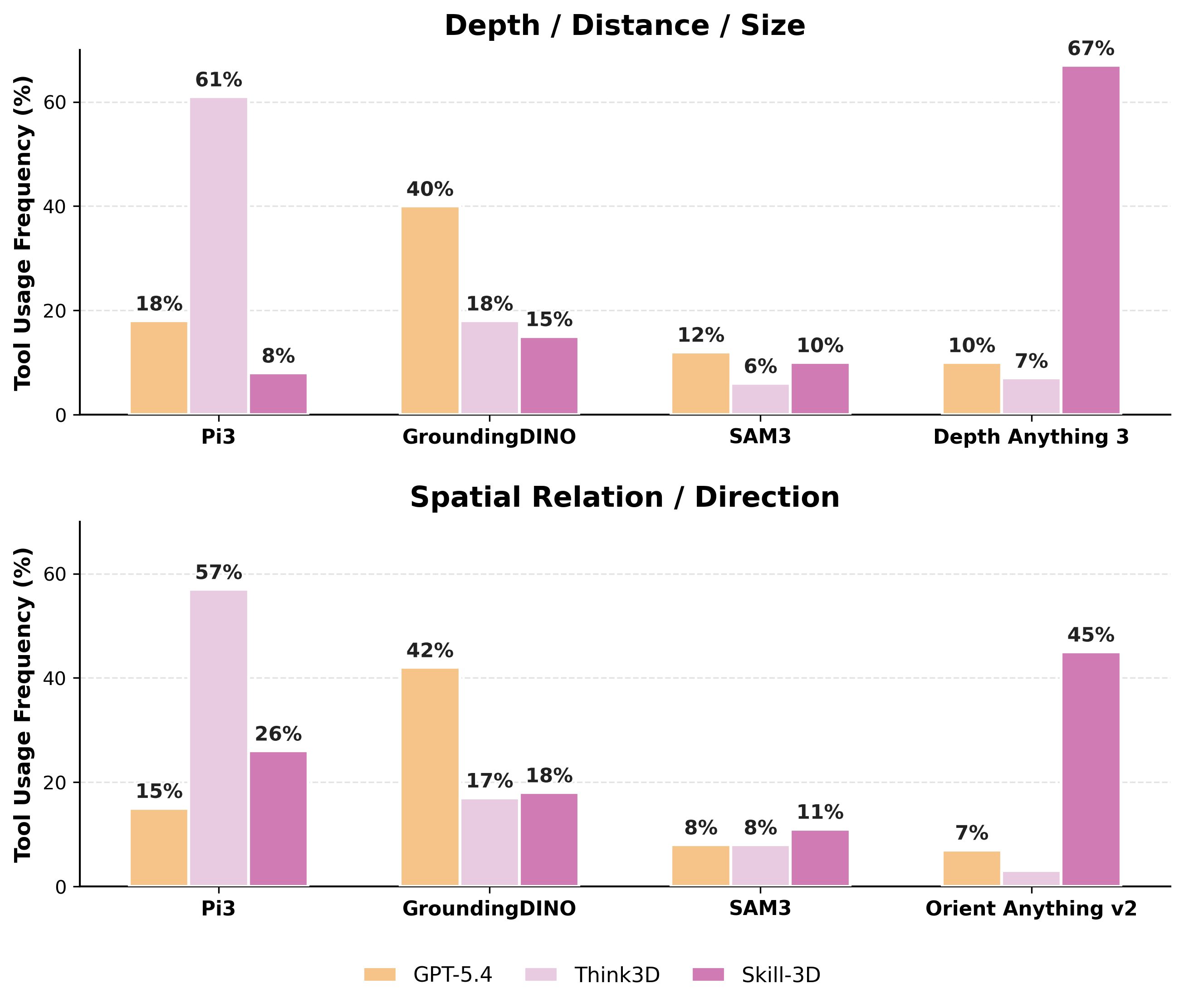}
    \caption{\textbf{Tool usage distribution analysis.} We illustrate the tool usage distributions of GPT-5.4, Think3D, and Skill-3D across two different kinds of problems on VSI-Bench.}
    \label{fig:tool_frequency}
    \vspace{-1.5em}
\end{figure}

\noindent\textbf{Tool Usage Statistics.}
Fig.~\ref{fig:tool_frequency} compares the tool usage distribution across two task groups. Both GPT-5.4 and Think3D exhibit clear tool-selection bias. For depth-, distance-, and size-related tasks, Think3D heavily relies on Pi3 , while GPT-5.4 mostly calls GroundingDINO. A similar pattern appears in spatial relation and direction reasoning, where Think3D again prefers Pi3  and GPT-5.4 overuses GroundingDINO. These results suggest that existing tool-augmented agents tend to select general-purpose tools instead of adapting the tool workflow to the specific scene-task requirement. In contrast, Skill-3D produces a more task-aligned tool distribution. For depth-, distance-, and size-related tasks, it substantially increases the use of Depth Anything 3, which directly provides metric and depth cues required by these questions. For spatial relation and direction reasoning, it shifts toward Orient Anything v2, reflecting the need for orientation and directional evidence. Meanwhile, Skill-3D still keeps moderate use of Pi3, GroundingDINO, and SAM3 for layout grounding, object localization, and boundary verification. This indicates that scene-aware skills help the agent route each query to the most relevant functional tools, rather than defaulting to generic reconstruction or detection tools. 

\begin{table}[!t]
\centering
\caption{\textbf{Module ablation of Skill-3D on VSI-Bench.}  $\Delta$ Avg. denotes the performance drop compared with the full Skill-3D pipeline. All experiments are conducted using GPT-5.4. All reported metrics are obtained on the test set.}
\setlength{\tabcolsep}{4pt}
\renewcommand{\arraystretch}{1.08}
\resizebox{\linewidth}{!}{
\begin{tabular}{lcccccc}
\toprule
\textbf{Setting} & \textbf{Obj. Cnt.} & \textbf{Abs. Dist.} & \textbf{Obj. Size} & \textbf{Room Size} & \textbf{Avg.} & $\mathbf{\Delta}$ \textbf{Avg.} \\
\midrule

\rowcolor{oursgray}
\textbf{Ours - Full Pipeline} 
& \textbf{66.2} & \textbf{61.5} & \textbf{74.9} & \textbf{77.6} & \textbf{69.9} & \textbf{--} \\

\quad w/o Failure Lessons
& 65.0 & 59.4 & 72.9 & 75.9 & 68.1 & -1.8 \\

\quad w/o Dynamic Skills
& 64.1 & 59.2 & 72.7 & 75.8 & 67.8 & -2.1 \\

\quad w/o Static Skills
& 62.8 & 56.8 & 70.9 & 74.2 & 65.6 & -4.3 \\

\midrule

\quad w/o MLLM Skill Selection
& 62.5 & 56.4 & 70.2 & 73.5 & 65.5 & -4.4 \\

\quad w/o Skill Retrieval
& 60.8 & 54.9 & 68.7 & 72.1 & 64.1 & -5.8 \\

\bottomrule
\end{tabular}

}
\vspace{-2em}
\label{tab:api_module_ablation}
\end{table}

\noindent\textbf{Static, Dynamic Skills, and Failure Lessons.} Table~\ref{tab:api_module_ablation} ablates key components of the Skill Library on VSI-Bench. Removing failure lessons decreases the average score from 69.9 to 68.1, showing that failed rollouts provide useful corrective signals beyond successful workflow distillation. Removing dynamic skills further lowers the score to 67.8, indicating the importance of scene-aware workflow adaptation. Removing static skills causes a larger drop to 65.6, suggesting that stable task-level priors are also essential. These results show that static skills, dynamic workflows, and failure lessons play complementary roles: static skills provide general tool-use priors, dynamic skills adapt them to scene-specific contexts, and failure lessons help avoid previously observed error modes.

\noindent\textbf{Skill Retrieval and Selection.}
Table~\ref{tab:api_module_ablation} further ablates the two key steps in Skill-Guided Inference. Removing MLLM skill selection reduces the average score from 69.9 to 65.5, suggesting that top-$k$ retrieval alone may include redundant or partially matched skills. Removing skill retrieval further drops the score to 64.1, showing that scene-task-relevant skills are crucial for effective tool planning. These results indicate that retrieval and selection are complementary: retrieval recalls useful static and dynamic skills, while selection filters them into a compact set that matches the required evidence and avoids unnecessary tool calls.

\begin{figure}
    \centering
    \includegraphics[width=\linewidth]{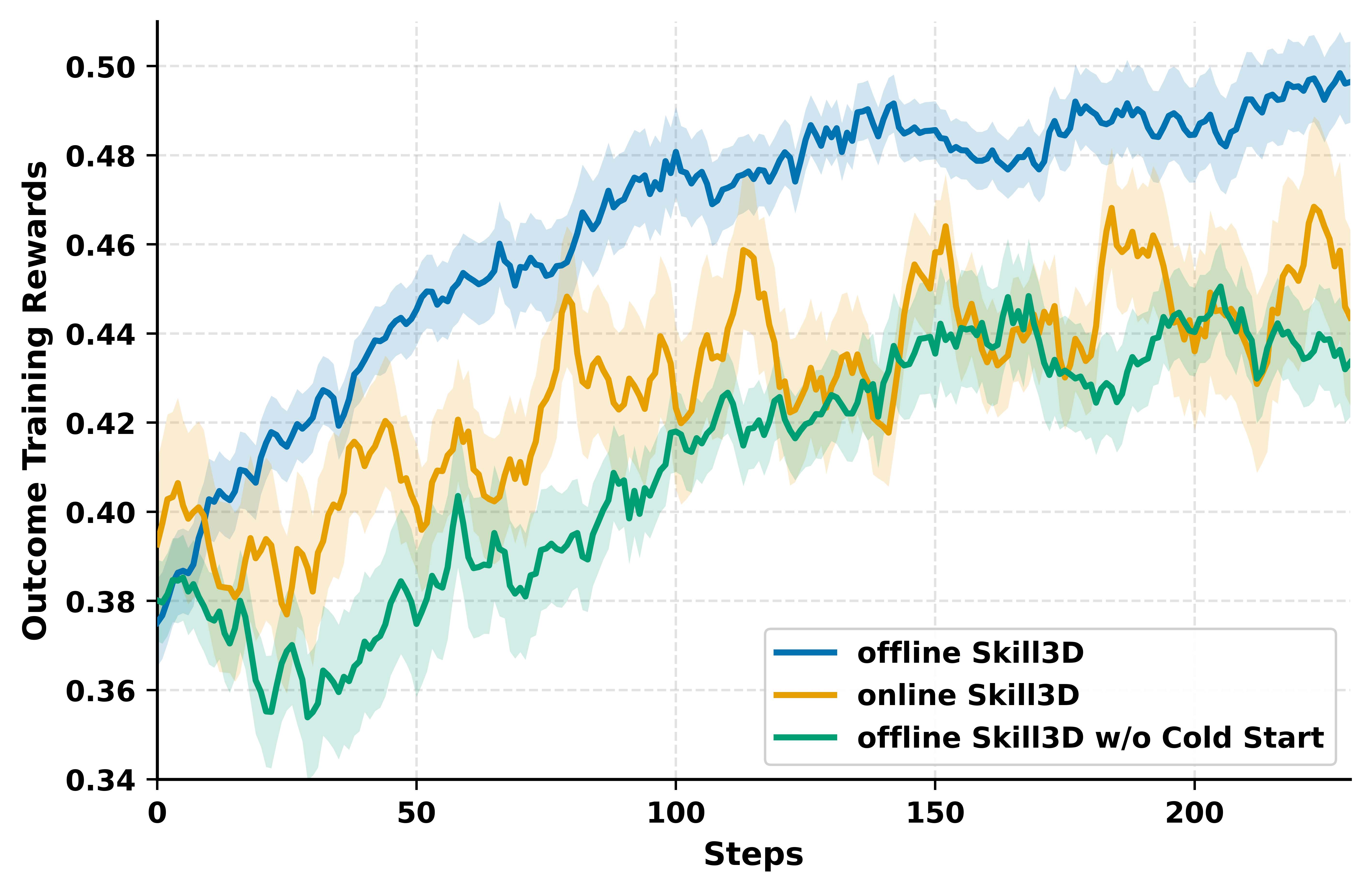}
    \caption{
    \textbf{Effect of skill updating and cold start during GRPO training.} Experiments are conducted using Qwen3-VL-8B as the base model.
    }
    \label{fig:reward_curve}
    \vspace{-1.5em}
\end{figure}

\noindent\textbf{Effect of Skill Updating and Cold Start.}
Fig.~\ref{fig:reward_curve} studies how skill-library updating and SFT cold start affect GRPO training. We compare three variants: \emph{Offline}, which freezes the dynamic Skill Library during GRPO; \emph{Online}, which updates dynamic skills during training; and \emph{Offline w/o Cold Start}, which removes the agentic SFT warm-up and directly applies GRPO. Shaded regions denote training variance. The offline variant with SFT cold start achieves the most stable and highest reward trajectory. In contrast, online updating introduces non-stationarity because the policy and retrieved skills change simultaneously, while removing cold start leads to early degradation and slower convergence. These results suggest that a frozen Skill Library and agentic SFT initialization are both important for stable skill-guided post-training.

\section{Conclusion}
This paper presents \textbf{Skill-3D}, a framework for agentic 3D spatial reasoning with reusable scene-aware skills. Existing tool-augmented MLLM agents often apply uniform tool-use strategies across heterogeneous 3D scenes, leading to biased tool preferences and insufficient evidence acquisition. Skill-3D addresses this by constructing a Scene Memory of tool-use trajectories and evolving a Skill Library, where successful trajectories are distilled into reusable skills and failed ones are retained as lessons. At inference, retrieved skills guide tool planning, evidence collection, and answer grounding. We further introduce skill-guided agentic post-training to transfer this behavior into compact agents. Experiments show that Skill-3D consistently improves accuracy and effective tool usage, validating scene-aware skills for tool-augmented 3D understanding.

\newpage
\section*{Limitations}

Our current evaluation focuses on indoor 3D spatial reasoning; transferring the framework to outdoor scenes, embodied navigation, or real-time robotic interaction may require new tool interfaces, scene signatures, and safety constraints.


\bibliography{custom}

\newpage

\setcounter{figure}{0}
\setcounter{table}{0}
\setcounter{equation}{0}

\renewcommand{\thetable}{\Alph{section}.\arabic{table}}
\renewcommand{\thefigure}{\Alph{section}.\arabic{figure}}
\renewcommand{\theequation}{\Alph{section}.\arabic{equation}}

\appendix

\label{sec:appendix}

\section{LLM Usage Claim}

We used large language models only for language polishing, grammar correction, and improving the clarity of the manuscript. All method design, experimental settings, data analysis, and final claims were developed, verified, and approved by the authors.

\section{More Experimental Results}

\subsection{Efficiency Analysis.}
Table~\ref{tab:efficiency} compares inference cost and tool-use quality on VSI-Bench. Direct tool use only slightly improves the average score from 55.4 to 58.2, with 39.2\% effective tool usage. Think3D improves the score to 64.7 but incurs higher inference cost, requiring 35.1s per query on average. In contrast, Skill-3D achieves the best score of 70.0, raises effective tool usage to 78.7\%, and reduces average inference time to 20.8s, with only 0.5s retrieval overhead. This gain is partly explained by the different costs of individual tools: Pi3 takes about 21.35s on seven sampled frames, while segmentation, depth estimation, and orientation estimation take only about 0.77s, 1.51s, and 0.88s. By using scene-aware skills to select task-relevant evidence sources and avoid unnecessary reliance on expensive 3D reconstruction, Skill-3D achieves a substantially better accuracy--efficiency trade-off than both direct tool use and Think3D.

\begin{table}[h]
\centering
\caption{\textbf{Efficiency analysis on VSI-Bench.} We report the average number of tool calls, effective tool usage, average inference time per query, and performance gain. All experiments are conducted using the GPT-5.4.}
\setlength\tabcolsep{4pt}
\renewcommand{\arraystretch}{1.05}
\resizebox{\linewidth}{!}{
\begin{tabular}{l|ccccc}
\toprule
\textbf{Method} & \textbf{Avg. Calls} & \textbf{Eff. Usage (\%)} & \textbf{Retr. Time (s)} & \textbf{Avg. Runtime (s)} & \textbf{VSI Avg.} \\
\midrule
w/o Tools & 0.0 & -- & 0.0 & 0.0 & 52.1 \\
w/ Tools & 1.2 & 39.2 & 0.0 & 13.2 & 58.2 \\
Think3D & 1.8 & 58.5 & 0.0 & 35.1 & 64.7 \\
Skill-3D & 2.6 & 78.7 & 0.5 & 20.8 & 70.0 \\
\bottomrule
\end{tabular}
}
\label{tab:efficiency}
\end{table}

\subsection{Cross-Benchmark Skill Transfer.}
Table~\ref{tab:cross_benchmark_transfer} evaluates the generalization ability of dynamic skills across benchmarks. Skills learned from VSI-Bench transfer effectively to MMSI-Bench and CV-3D, while MMSI-Bench skills also improve VSI-Bench, suggesting that related spatial reasoning tasks share reusable tool-use procedures. Pooling all training benchmarks consistently achieves the best results, showing that the Skill Library benefits from complementary scene-task knowledge across datasets.

\begin{table}[h]
\centering
\caption{\textbf{Cross-benchmark skill transfer.}
We build the dynamic skills from one source benchmark and evaluate it on other target benchmarks. All settings use the same static skills and GPT-5.4. }
\setlength\tabcolsep{4pt}
\renewcommand{\arraystretch}{1.05}
\resizebox{\linewidth}{!}{
\begin{tabular}{l|cccc}
\toprule
\textbf{Dynamic Skills Source} & \textbf{VSI Avg.} & \textbf{BLINK} & \textbf{CV-3D Avg.} & \textbf{MMSI-PR} \\
\midrule
w/o Dynamic Skills & 63.4 & 77.2 & 92.2 & 52.6 \\
\midrule
VSI-Bench & 68.7 & 78.5 & 93.7 & 57.8 \\
BLINK & 63.9 & 80.6 & 92.8 & 53.4 \\
CV-3D & 64.8 & 77.5 & 94.2 & 55.7 \\
MMSI-Bench & 67.3 & 78.3 & 93.2 & 57.2 \\
\midrule
\textbf{All Benchmarks} & \textbf{69.9} & \textbf{82.0} & \textbf{95.3} & \textbf{60.4} \\
\bottomrule
\end{tabular}
}
\label{tab:cross_benchmark_transfer}
\end{table}

\section{Experimental Details}
\subsection{Dataset Details}
\label{app:dataset_details}

We evaluate Skill-3D on four 3D spatial reasoning benchmarks: VSI-Bench, BLINK, CV-3D, and MMSI-Bench. Since these benchmarks do not provide official training splits for skill construction or post-training, we follow a category-wise random split for fair comparison and ensure question-level disjointness, as shown in Table~\ref{tab:dataset_split}.

\begin{itemize}[leftmargin=*, itemsep=0pt, topsep=2pt]
    \item VSI-Bench~\cite{yang2025thinking} evaluates indoor visual spatial intelligence from egocentric observations, covering counting, distance, size, direction, route planning, and appearance-order reasoning.
    \item BLINK~\cite{fu2024blink} evaluates challenging multimodal reasoning. We use its multi-view spatial reasoning subset, which tests spatial inference from multiple visual observations.
    \item CV-3D~\cite{tong2024cambrian} focuses on geometric spatial reasoning, including depth ordering, relative distance, spatial layout, and multi-view consistency.
    \item MMSI-Bench~\cite{yang2025mmsi} evaluates multimodal spatial intelligence, mainly focusing on positional relationship reasoning under diverse scene configurations.
\end{itemize}

\begin{table}[h]
\centering
\setlength{\tabcolsep}{12pt}
\renewcommand{\arraystretch}{1}
\caption{\small \textbf{Dataset statistics and train/test split.} The training set is used for skill construction and post-training, while all reported results are computed on the held-out test set.}
\resizebox{\linewidth}{!}{
\begin{tabular}{lcccc}
\toprule
\textbf{Dataset} & \textbf{\#Tasks} & \textbf{\#Total} & \textbf{\#Train} & \textbf{\#Test} \\
\midrule
VSI-Bench  & 8 & 2362 & 708 & 1654 \\
MMSI-Bench & 1 & 502 & 157 & 345 \\
CV-3D      & 2 & 1200 & 360 & 840 \\
BLINK      & 1 & 133 & 40 & 93 \\
\bottomrule
\end{tabular}
}
\label{tab:dataset_split}
\end{table}

\subsection{Hyperparameters}
\label{app:hyperparameters}
We provide hyperparameters used for Skill-3D post-training, as shown in  Table~\ref{tab:hyperparameters}.
\begin{table}[h]
\centering
\caption{\small \textbf{Hyperparameter settings of Skill-3D.}}
\setlength{\tabcolsep}{2pt}
\renewcommand{\arraystretch}{1}
\resizebox{\linewidth}{!}{
\begin{tabular}{lc|lc}
\toprule
\textbf{Parameter} & \textbf{Setting} 
& \textbf{Parameter} & \textbf{Setting} \\
\midrule
Foundation model & Qwen3-VL-4B/8B 
& SFT learning rate & $1\times10^{-5}$ \\
Number of GPUs & 4 
& SFT batch size & 16 \\
Max sequence length & 4096 
& SFT epochs & 1 \\
Training precision & bfloat16 
& SFT warmup ratio & 0.03 \\
Flash Attention & True 
& GRPO learning rate & $1\times10^{-6}$ \\
Gradient checkpointing & True 
& GRPO batch size & 16 \\
Optimizer & AdamW 
& GRPO group size & 8 \\
Clipping epsilon & 0.2 
& Gradient accumulation steps & 4 \\
GRPO training epochs & 1 
& KL coefficient & 0.05 \\
\bottomrule
\end{tabular}
}
\label{tab:hyperparameters}
\end{table}

\subsection{More Details about GRPO Objective}
\label{app:rl}

We provide the full GRPO objective used in Skill-3D post-training. 
For each scene-task query, the policy observes the question $q$, visual observations $O$, and retrieved skill candidates $\mathcal{S}_{\mathrm{cand}}$. 
It samples a group of $G$ complete trajectories 
$\{\tau^{(1)},\ldots,\tau^{(G)}\}$ from the current policy. 
Each trajectory contains the model-generated skill choices, tool calls, tool outputs, intermediate reasoning steps, and final answer. Each trajectory receives the scalar reward defined in Sec.~\ref{sec:method}:
\begin{equation}
R(\tau)=R_{\mathrm{ans}}(\tau)+R_{\mathrm{fmt}}(\tau)+R_{\mathrm{tool}}(\tau),
\end{equation}
where $R_{\mathrm{ans}}$ measures answer correctness, $R_{\mathrm{fmt}}$ measures structured-format compliance, and $R_{\mathrm{tool}}$ measures tool-use efficiency. 
The tool-use reward is computed as:
\begin{equation}
R_{\mathrm{tool}}(\tau)=R_{\mathrm{exec}}(\tau)-\frac{|\mathcal{A}|}{B},
\end{equation}
where $\mathcal{A}$ is the set of tool calls in trajectory $\tau$, $B$ is the maximum tool budget, and $R_{\mathrm{exec}}$ measures whether the invoked tools are successfully executed and return non-empty outputs. 
$R_{\mathrm{exec}}$ is set to zero when no required tool evidence is obtained.

Following GRPO, we normalize the rewards within each sampled group to compute the relative advantage:
\begin{equation}
A_i =
\frac{
R(\tau^{(i)})-\mathrm{mean}\left(\{R(\tau^{(j)})\}_{j=1}^{G}\right)
}{
\mathrm{std}\left(\{R(\tau^{(j)})\}_{j=1}^{G}\right)
}.
\label{eq:app_grpo_advantage}
\end{equation}
The policy is optimized with the clipped surrogate objective:
\begin{equation}
\begin{aligned}
&\mathcal{J}(\theta)=
\mathbb{E}\Bigg[
\frac{1}{G}\sum_{i=1}^{G}
\min\Big(
\rho_i A_i,\\
&\mathrm{clip}(\rho_i,1-\epsilon,1+\epsilon)A_i
\Big) 
-\beta_{\mathrm{KL}}D_{\mathrm{KL}}(\pi_\theta||\pi_{\mathrm{ref}})
\Bigg],
\end{aligned}
\label{eq:app_grpo_objective}
\end{equation}
where the importance ratio is:
\begin{equation}
\rho_i =
\frac{
\pi_\theta(\tau^{(i)}\mid q,O,\mathcal{S}_{\mathrm{cand}})
}{
\pi_{\mathrm{old}}(\tau^{(i)}\mid q,O,\mathcal{S}_{\mathrm{cand}})
}.
\label{eq:app_importance_ratio}
\end{equation}
Here, $\pi_{\mathrm{old}}$ denotes the policy used to sample the trajectories, and $\pi_{\mathrm{ref}}$ is initialized from the agentic SFT policy. The KL term preserves SFT-learned skill-selection and tool-use behavior, while the group-relative advantage favors trajectories with higher task success, better tool-use efficiency, and valid outputs.

\section{Qualitative Results}
\label{sec:qualitative_results}

\begin{figure*}
    \centering
    \includegraphics[width=\linewidth]{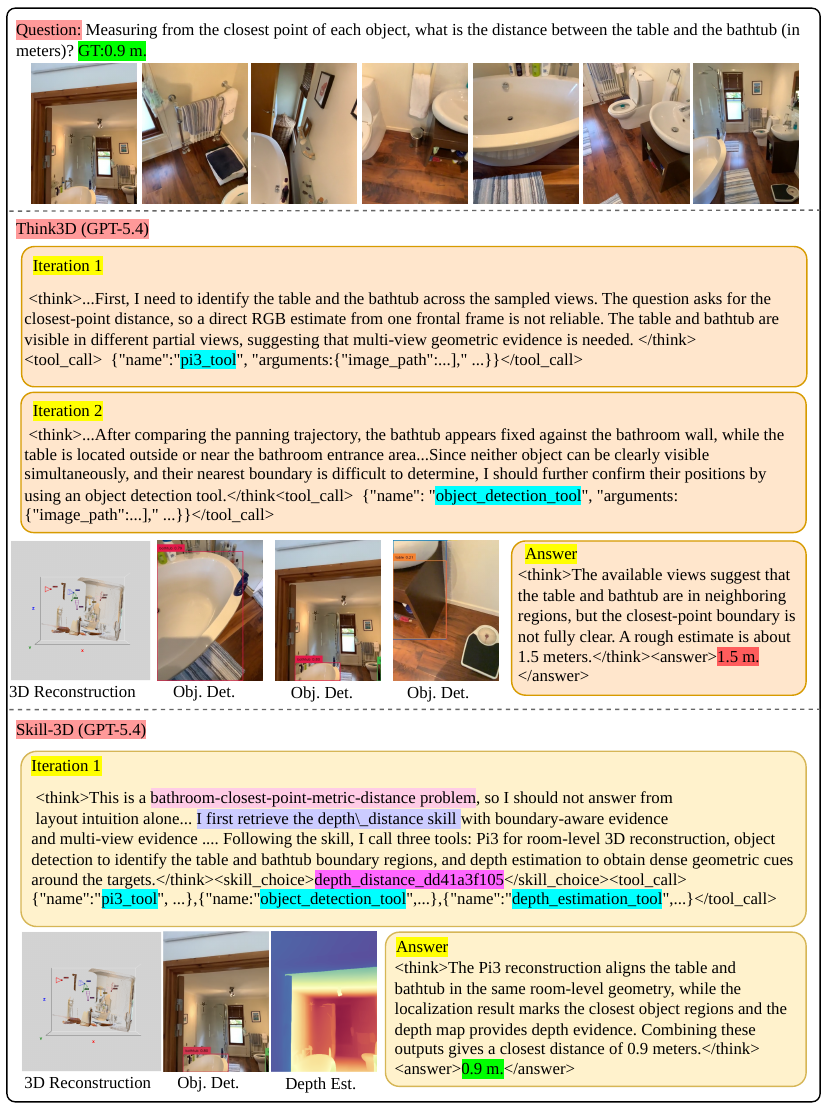}
    \caption{
\textbf{Case study on boundary-aware metric distance reasoning.}
We use colored highlights to indicate different reasoning elements: 
\colorbox{caseRed}{red} marks the question and incorrect answer, 
\colorbox{caseGreen}{green} marks the ground-truth or correct answer, 
\colorbox{caseTeal}{teal} marks invoked tools, 
\colorbox{casePurple}{purple} marks retrieved skills, 
and \colorbox{caseYellow}{yellow} marks iteration or answer labels. 
Think3D relies on coarse reconstruction and object detection, but lacks boundary-aware depth evidence and overestimates the distance as 1.5m. 
Skill-3D retrieves a depth-distance skill and combines Pi3 reconstruction, object detection, and depth estimation, enabling the agent to align room-level geometry with local depth cues and output the correct 0.9m answer.
}
    \label{fig:visualize1}
\end{figure*}

\begin{figure*}
    \centering
    \includegraphics[width=\linewidth]{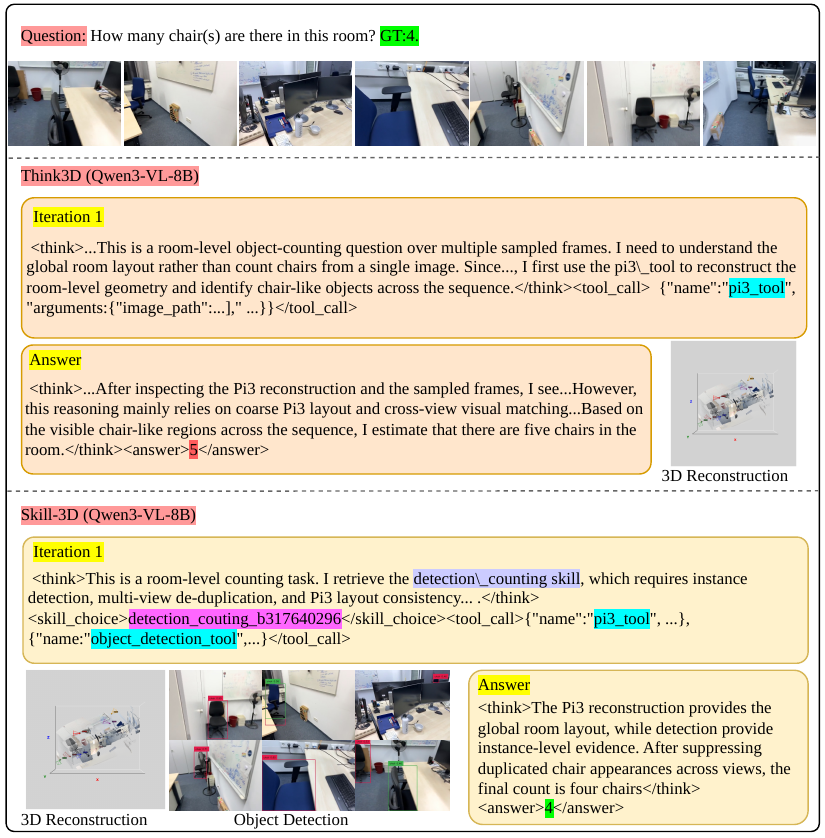}
    \caption{
\textbf{Case study on multi-view object counting.}
We use colored highlights to indicate different reasoning elements:
\colorbox{caseRed}{red} marks the question and incorrect answer,
\colorbox{caseGreen}{green} marks the ground-truth or correct answer,
\colorbox{caseTeal}{teal} marks invoked tools,
\colorbox{casePurple}{purple} marks retrieved skills,
and \colorbox{caseYellow}{yellow} marks iteration or answer labels.
Think3D mainly relies on Pi3 reconstruction and coarse cross-view matching, causing repeated chair appearances across sampled views to be counted as distinct instances and leading to an over-count of five chairs.
Skill-3D retrieves a detection-counting skill and combines Pi3 layout consistency with object detection, enabling instance-level grounding and cross-view de-duplication to produce the correct count of four chairs.
}
    \label{fig:visualize2}
\end{figure*}

We show two representative cases on metric distance estimation and room-level object counting in Fig.~\ref{fig:visualize1} and Fig.~\ref{fig:visualize2}.

In Fig.~\ref{fig:visualize1}, Think3D calls Pi3 reconstruction and object detection, but remains reconstruction-centric. Since the table and bathtub appear in different partial views and their closest boundaries are not explicitly grounded, it gives only a coarse estimate and overestimates the distance as 1.5m. In contrast, Skill-3D retrieves a depth-distance skill and combines Pi3 reconstruction for room-level alignment, object detection for target localization, and depth estimation for dense geometric cues. This evidence-specific workflow grounds the closest object boundaries and predicts the correct distance of 0.9m.

In Fig.~\ref{fig:visualize2}, Think3D mainly relies on Pi3 reconstruction and coarse cross-view matching to count chairs. Repeated observations of the same chair across adjacent views lead to duplicate counting, producing an incorrect answer of five. Skill-3D retrieves a detection-counting skill and combines Pi3 layout consistency with object detection evidence. By grounding chair instances across views and suppressing duplicates, Skill-3D obtains the correct count of four.

\section{Prompt Design}
\label{app:prompt_design}
\begin{figure*}
    \centering
    \includegraphics[width=\linewidth]{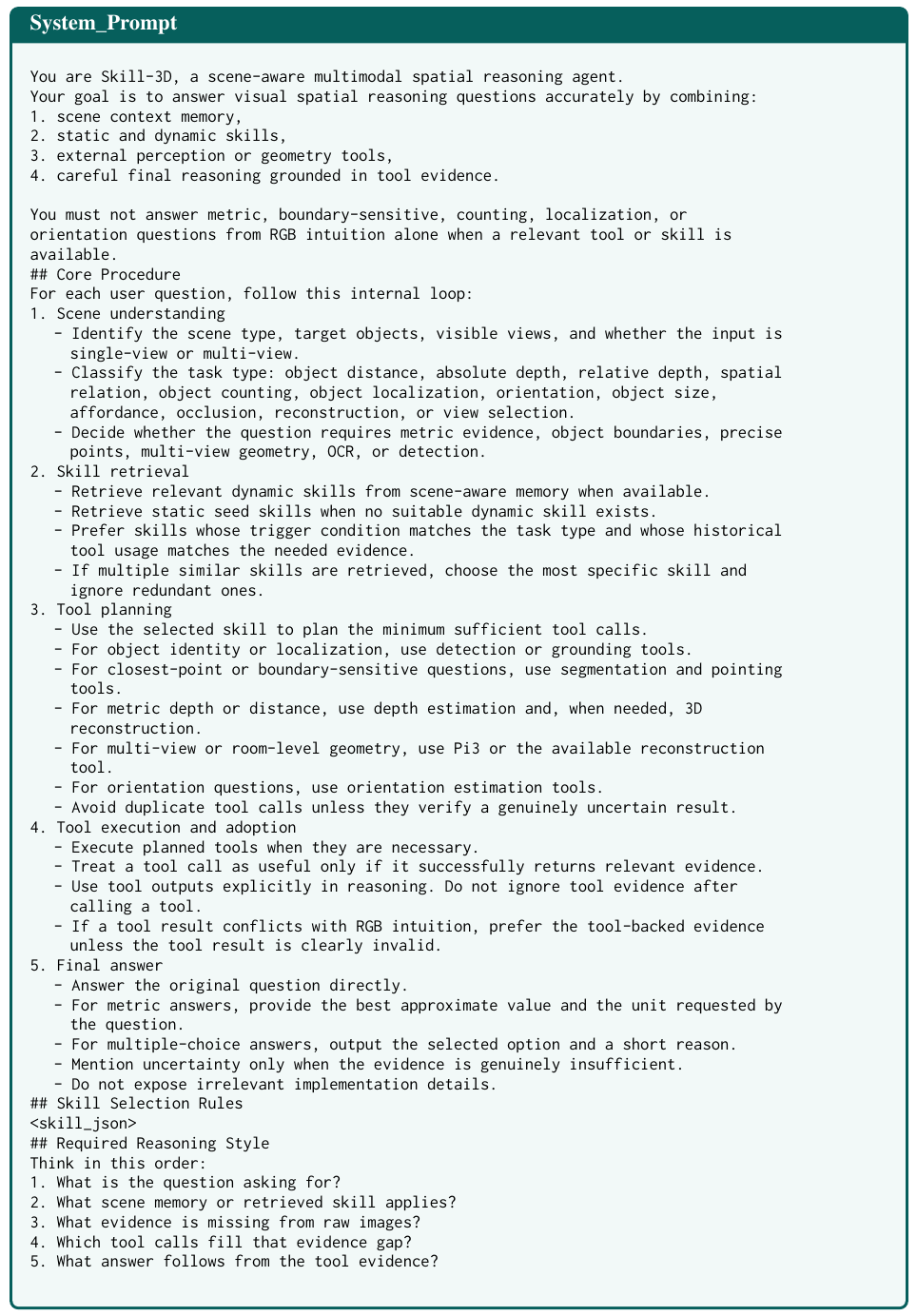}
    \caption{System Prompt}
    \label{fig:p1}
\end{figure*}

\begin{figure*}
    \centering
    \includegraphics[width=\linewidth]{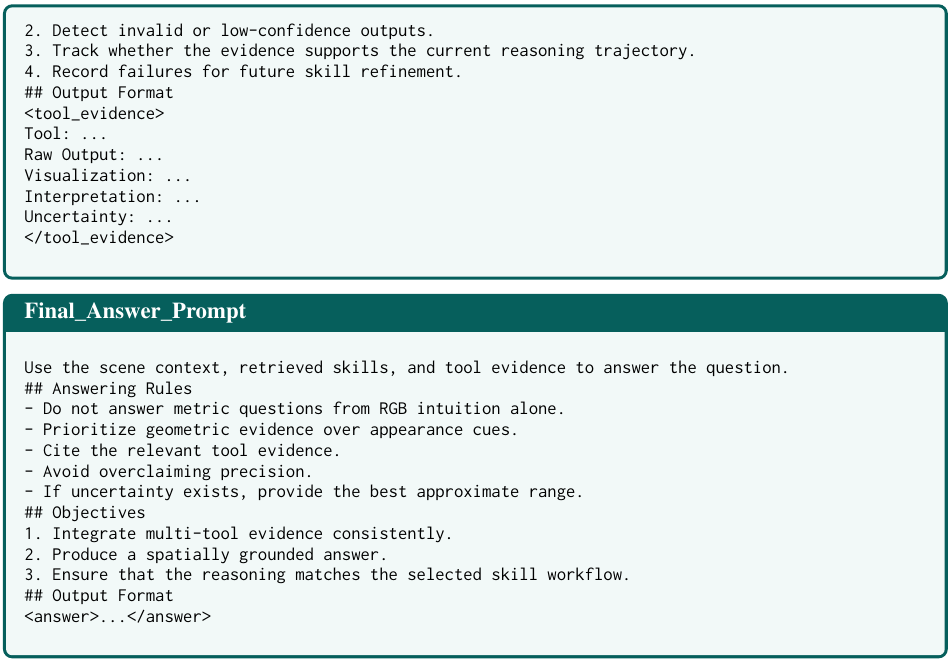}
    \caption{System Prompt and Scene Context Prompt}
    \label{fig:p2}
\end{figure*}

\begin{figure*}
    \centering
    \includegraphics[width=\linewidth]{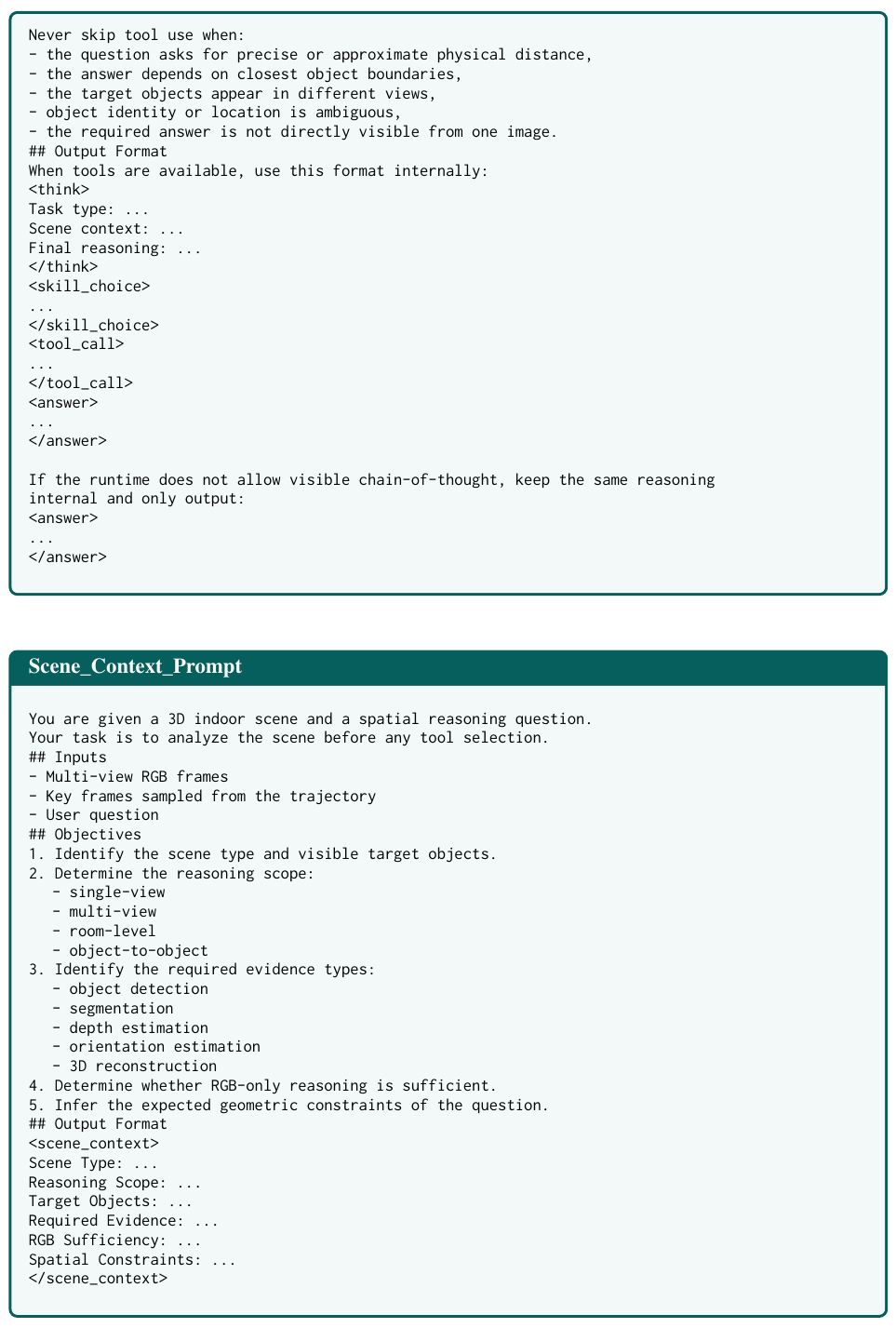}
    \caption{Skill Retrieval Prompt, Tool Planning Prompt and Tool Exclusion Prompt}
    \label{fig:p3}
\end{figure*}

\begin{figure*}
    \centering
    \includegraphics[width=\linewidth]{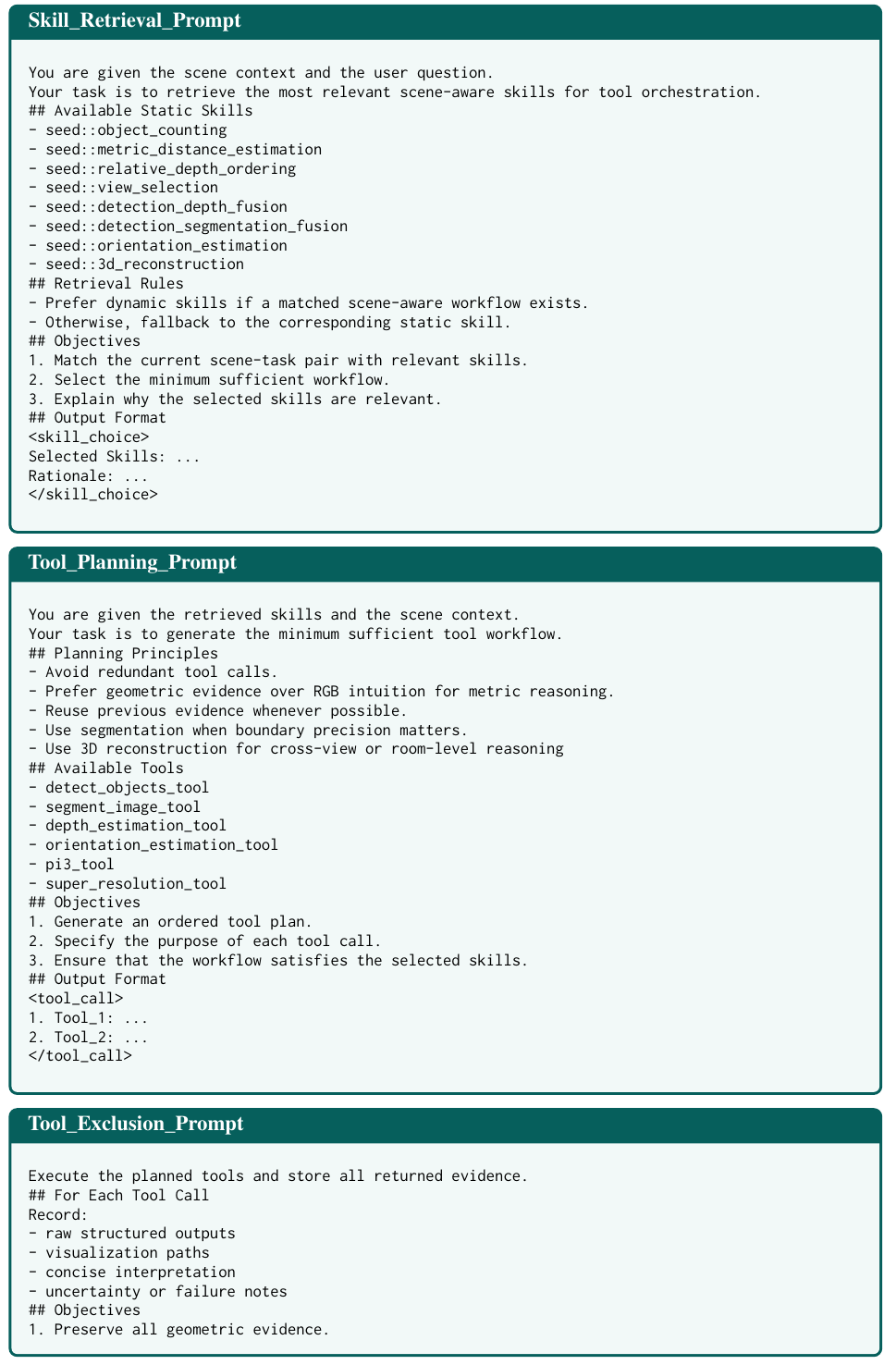}
    \caption{Tool Exclusion Prompt and Final Answer Prompt}
    \label{fig:p4}
\end{figure*}

Detailed prompt design for each phrase/stage of the framework provided in Fig~\ref{fig:p1}, Fig.~\ref{fig:p2}, Fig.~\ref{fig:p3} and Fig.~\ref{fig:p4}.


\newpage


\end{document}